\documentclass[12pt,twoside]{article}
\usepackage{latexsym,amsmath,amssymb}
\def\eqvsp{}  \newdimen\paravsp  \paravsp=1.3ex
\def\,{\mskip 3mu} \def\>{\mskip 4mu plus 2mu minus 4mu} \def\;{\mskip 5mu plus 5mu} \def\!{\mskip-3mu}
\def\dispmuskip{\thinmuskip= 3mu plus 0mu minus 2mu \medmuskip=  4mu plus 2mu minus 2mu \thickmuskip=5mu plus 5mu minus 2mu}
\def\textmuskip{\thinmuskip= 0mu                    \medmuskip=  1mu plus 1mu minus 1mu \thickmuskip=2mu plus 3mu minus 1mu}
\textmuskip
\def\beq{\eqvsp\dispmuskip\begin{equation}}    \def\eeq{\eqvsp\end{equation}\textmuskip}
\def\beqn{\eqvsp\dispmuskip\begin{displaymath}}\def\eeqn{\eqvsp\end{displaymath}\textmuskip}
\def\bqa{\eqvsp\dispmuskip\begin{eqnarray}}    \def\eqa{\eqvsp\end{eqnarray}\textmuskip}
\def\bqan{\eqvsp\dispmuskip\begin{eqnarray*}}  \def\eqan{\eqvsp\end{eqnarray*}\textmuskip}

\newenvironment{keywords}{\centerline{\bf\small
Keywords}\begin{quote}\small}{\par\end{quote}\vskip 1ex}
\def\paradot#1{\vspace{\paravsp plus 0.5\paravsp minus 0.5\paravsp}\noindent{\bf\boldmath{#1.}}}
\def\paranodot#1{\vspace{\paravsp plus 0.5\paravsp minus 0.5\paravsp}\noindent{\bf\boldmath{#1}}}
\def\req#1{(\ref{#1})}

\def\epstr{\epsilon}                    
\def\nq{\hspace{-1em}}
\def\fr#1#2{{\textstyle{#1\over#2}}}
\def\SetR{I\!\!R}
\def\SetN{I\!\!N}
\def\qmbox#1{{\quad\mbox{#1}\quad}}
\def\P{{\rm P}}                         
\def\v{\vec}
\def\v{\boldsymbol}

\def\a{\alpha}
\def\g{\gamma}
\def\t{\theta}
\def\A{{\cal A}}
\def\O{{\cal O}}
\def\R{{\cal R}}
\def\S{{\cal S}}
\def\H{{\cal H}}
\def\X{{\cal X}}
\def\T{{\cal T}}
\def\Agent{\text{Agent}}
\def\Env{\text{Env}}
\def\p{{\scriptscriptstyle+}}
\def\vi{{\scriptscriptstyle\bullet}}
\def\CL{\text{CL}}
\def\CLN{\text{CL}}
\def\Cost{\text{Cost}}

\begin{document}

\title{\vspace{-4ex}
\vskip 2mm\bf\Large\hrule height5pt \vskip 4mm
Feature Reinforcement Learning: \\ Part I. Unstructured MDPs
\vskip 4mm \hrule height2pt}
\author{{\bf Marcus Hutter%
\thanks{A shorter version appeared in the proceedings of the AGI 2009 conference \cite{Hutter:09phimdp}.}}\\[3mm]
\normalsize RSISE$\,$@$\,$ANU and SML$\,$@$\,$NICTA \\
\normalsize Canberra, ACT, 0200, Australia \\
\normalsize \texttt{marcus@hutter1.net \ \  www.hutter1.net}
}

\date{9 June 2009}
\maketitle
\vspace*{-5ex}

\begin{abstract}
General-purpose, intelligent, learning agents cycle through
sequences of observations, actions, and rewards that are complex,
uncertain, unknown, and non-Markovian. On the other hand,
reinforcement learning is well-developed for small finite state
Markov decision processes (MDPs). Up to now, extracting the right
state representations out of bare observations, that is, reducing
the general agent setup to the MDP framework, is an art that
involves significant effort by designers. The primary goal of this
work is to automate the reduction process and thereby significantly
expand the scope of many existing reinforcement learning algorithms
and the agents that employ them. Before we can think of mechanizing
this search for suitable MDPs, we need a formal objective criterion.
The main contribution of this article is to develop such a
criterion. I also integrate the various parts into one learning
algorithm. Extensions to more realistic dynamic Bayesian networks
are developed in Part II \cite{Hutter:09phidbnx}. The role of POMDPs
is also considered there.
\def\contentsname{\centering\normalsize Contents}
{\parskip=-2.7ex\tableofcontents}
\end{abstract}\vspace{-2ex}


\newpage

\begin{quote}\it
``Approximations, after all, may be made in two places -
in the construction of the model and in the solution of the
associated equations. It is not at all clear which yields
a more judicious approximation.'' \par
\hfill --- {\sl Richard Bellman (1961)}
\end{quote}

\section{Introduction}\label{secIntro}

\paradot{Background \& motivation}
Artificial General Intelligence (AGI) is concerned with designing
agents that perform well in a wide range of environments
\cite{Goertzel:07,Hutter:07iorx}.
%
Among the well-established ``narrow'' Artificial Intelligence (AI)
approaches \cite{Russell:03}, arguably Reinforcement Learning (RL)
\cite{Sutton:98} pursues most directly the same goal. RL considers
the general agent-environment setup in which an agent interacts with
an environment (acts and observes in cycles) and receives
(occasional) rewards. The agent's objective is to collect as much
reward as possible. Most if not all AI problems can be formulated in
this framework. Since the future is generally unknown and uncertain,
the agent needs to learn a model of the environment based on past
experience, which allows to predict future rewards and use this to
maximize expected long-term reward.

The simplest interesting environmental class consists of finite
state fully observable Markov Decision Processes (MDPs)
\cite{Puterman:94,Sutton:98}, which is reasonably well understood.
%
Extensions to continuous states with (non)linear function
approximation \cite{Sutton:98,Gordon:99}, partial observability
(POMDP) \cite{Kaelbling:98,Ross:08pomdp}, structured MDPs (DBNs)
\cite{Strehl:07}, and others have been considered, but the
algorithms are much more brittle.

A way to tackle complex real-world problems is to reduce them to
finite MDPs which we know how to deal with efficiently. This
approach leaves a lot of work to the designer, namely to extract the
right state representation (``features'') out of the bare
observations in the initial (formal or informal) problem
description. Even if {\em potentially} useful representations have
been found, it is usually not clear which ones will turn out to be
better, except in situations where we already know a perfect model.
Think of a mobile robot equipped with a camera plunged into an
unknown environment. While we can imagine which image features will
potentially be useful, we cannot know in advance which ones will
actually be useful.

\paradot{Main contribution}
The primary goal of this paper is to develop and investigate a
method that {\em automatically} selects those features that are
necessary and sufficient for {\em reducing} a complex real-world
problem to a computationally tractable MDP.

Formally, we consider maps $\Phi$ from the past
observation-reward-action history $h$ of the agent to an MDP state.
Histories not worth being distinguished are mapped to the same
state, i.e.\ $\Phi^{-1}$ induces a partition on the set of
histories. We call this model $\Phi$MDP. A state may be simply an
abstract label of the partition, but more often is itself a
structured object like a discrete vector. Each vector component
describes one feature of the history
\cite{Hutter:09phidbn,Hutter:09phidbnx}. For example, the state may
be a 3-vector containing (shape,color,size) of the object a robot
tracks.
For this reason, we call the {\em reduction}, {\em Feature RL},
although in this Part I only the simpler unstructured case is
considered.

$\Phi$ maps the agent's experience over time into a sequence of MDP
states.
Rather than informally constructing $\Phi$ by hand, our goal is to
develop a formal objective criterion $\Cost(\Phi|h)$ for {\em
evaluating} different reductions $\Phi$. Obviously, at any point in
time, if we want the criterion to be effective it can only depend on
the agent's past experience $h$ and possibly generic background
knowledge.
The ``Cost'' of $\Phi$ shall be small iff it leads to a ``good'' MDP
representation. The establishment of such a criterion transforms
the, in general, ill-defined RL problem to a formal optimization
problem (minimizing Cost) for which efficient algorithms need to be
developed. Another important question is which problems {\em can}
profitably be reduced to MDPs
\cite{Hutter:09phidbn,Hutter:09phidbnx}.

The real world does not conform itself to nice models: Reality is a
non-ergodic partially observable uncertain unknown environment in
which acquiring experience can be expensive. So we should exploit
the data (past experience) at hand as well as possible, cannot
generate virtual samples since the model is not given (need to be
learned itself), and there is no reset-option.
No criterion for this general setup exists. Of course, there is
previous work which is in one or another way related to $\Phi$MDP.

\paradot{$\Phi$MDP in perspective}
As partly detailed later, the suggested $\Phi$MDP model has
interesting connections to many important ideas and approaches in RL
and beyond:
\begin{itemize}\parskip=0ex\parsep=0ex\itemsep=0ex
\item $\Phi$MDP side-steps the open problem of learning POMDPs \cite{Kaelbling:98}, %
\item Unlike Bayesian RL algorithms \cite{Dearden:99,Duff:02,Poupart:06,Ross:08bayes}, $\Phi$MDP avoids learning a (complete stochastic) observation model, %
\item $\Phi$MDP is a scaled-down practical instantiation of AIXI \cite{Hutter:04uaibook,Hutter:07aixigentle}, %
\item $\Phi$MDP extends the idea of state-aggregation from planning (based on bi-simulation metrics \cite{Givan:03}) to RL (based on information), %
\item $\Phi$MDP generalizes U-Tree \cite{McCallum:96} to arbitrary features, %
\item $\Phi$MDP extends model selection criteria to general RL problems \cite{Gruenwald:07book}, %
\item $\Phi$MDP is an alternative to PSRs \cite{Singh:03} for which proper learning algorithms have yet to be developed, %
\item $\Phi$MDP extends feature selection from supervised learning to RL \cite{Guyon:03}.
\end{itemize}
Learning in agents via rewards is a much more demanding task than
``classical'' machine learning on independently and identically
distributed (i.i.d.) data, largely due to the temporal credit
assignment and exploration problem. Nevertheless, RL (and the
closely related adaptive control theory in engineering) has been
applied (often unrivaled) to a variety of real-world problems,
occasionally with stunning success (Backgammon, Checkers,
\cite[Chp.11]{Sutton:98}, helicopter control \cite{Ng:04}).
$\Phi$MDP overcomes several of the limitations of the approaches in
the items above and thus broadens the applicability of RL.

$\Phi$MDP owes its general-purpose {\em learning} and {\em planning}
ability to its {\em information} and {\em complexity} theoretical
foundations. The implementation of $\Phi$MDP is based on
(specialized and general) {\em search} and {\em optimization}
algorithms used for finding good reductions $\Phi$.
%
Given that $\Phi$MDP aims at general AI problems, one may wonder about the
role of other aspects traditionally considered in AI
\cite{Russell:03}:
{\em knowledge representation} (KR) and {\em
logic} may be useful for representing complex reductions $\Phi(h)$.
Agent interface fields like {\em robotics}, computer {\em vision},
and natural {\em language} processing can speedup learning by
pre\&post-processing the raw observations and actions into more
structured formats. These representational and interface aspects
will only barely be discussed in this paper.
%
The following diagram illustrates $\Phi$MDP in perspective.

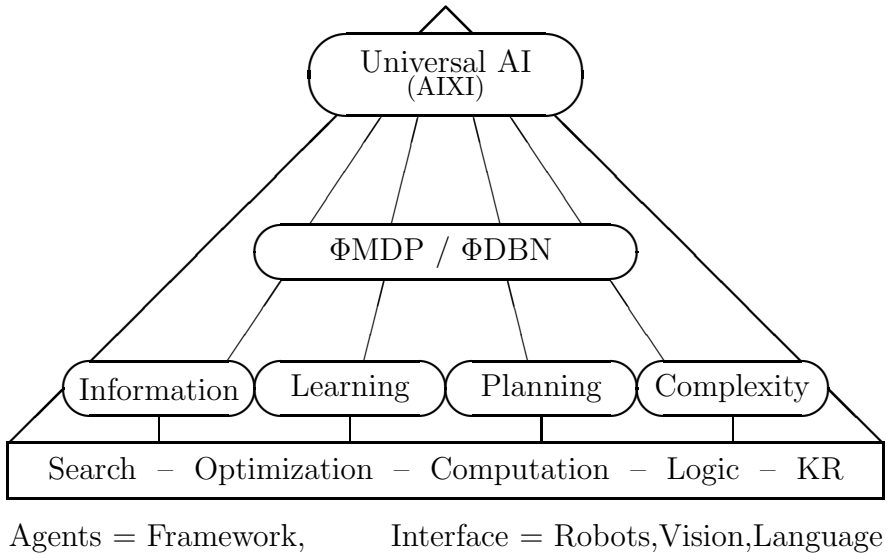
\begin{center}
\unitlength=2ex
\linethickness{0.4pt}
\begin{picture}(32,20)(0,0)
\thicklines
\put(16,17.5){\oval(10,3)\makebox(0,1)[cc]{Universal AI}\makebox(0,-1)[cc]{\footnotesize(AIXI)}}
\put(16,11){\oval(14,2)\makebox(0,0)[cc]{$\Phi$MDP / $\Phi$DBN }}
\put(5.5,6){\oval(7,2)\makebox(0,0)[cc]{Information}}
\put(12.5,6){\oval(7,2)\makebox(0,0)[cc]{Learning}}
\put(19.5,6){\oval(7,2)\makebox(0,0)[cc]{Planning}}
\put(26.5,6){\oval(7,2)\makebox(0,0)[cc]{Complexity}}
\put(0,2){\framebox(32,2)[cc]{Search ~--~ Optimization ~--~ Computation ~--~ Logic ~--~ KR}}
\put(0,4){\line(1,1){2}}\put(3,7){\line(1,1){9}}\put(15,19){\line(1,1){1}}
\put(32,4){\line(-1,1){2}}\put(29,7){\line(-1,1){9}}\put(17,19){\line(-1,1){1}}
\thinlines
\put(8,7){\line(2,3){2}}\put(11,12){\line(2,3){2.667}}
\put(24,7){\line(-2,3){2}}\put(21,12){\line(-2,3){2.667}}
\put(13,7){\line(1,4){0.75}}\put(14,12){\line(1,4){1}}
\put(19,7){\line(-1,4){0.75}}\put(18,12){\line(-1,4){1}}
\put(5.5,4){\line(0,1){1}}
\put(12.5,4){\line(0,1){1}}
\put(19.5,4){\line(0,1){1}}
\put(26.5,4){\line(0,1){1}}
\thicklines
\put(0,0){\makebox(16,2)[lb]{Agents = Framework,}}
\put(16,0){\makebox(16,2)[rb]{Interface = Robots,Vision,Language}}
\end{picture}
\end{center}

\paradot{Contents}
Section \ref{secPhiMDP} formalizes our $\Phi$MDP setup,
which consists of the agent model with a map $\Phi$ from
observation-reward-action histories to MDP states.
Section \ref{secCE} develops our core $\Phi$ selection principle,
which is illustrated in Section \ref{secTE} on a tiny example.
Section \ref{secCM} discusses general search algorithms for finding
(approximations of) the optimal $\Phi$, concretized
for context tree MDPs.
In Section \ref{secEE} I find the optimal action for
$\Phi$MDP, and present the overall algorithm.
Section \ref{secICF} improves the $\Phi$ selection criterion by
``integrating'' out the states.
Section \ref{secDisc} contains a brief discussion of $\Phi$MDP,
including relations to prior work, incremental algorithms, and an
outlook to more realistic {\em structured} MDPs (dynamic Bayesian
networks, $\Phi$DBN) treated in Part II.

Rather than leaving parts of $\Phi$MDP vague and unspecified, I
decided to give at the very least a simplistic concrete algorithm
for each building block, which may be assembled to one sound system
on which one can build on.

\paradot{Notation}
Throughout this article, $\log$ denotes the binary logarithm, %
$\epstr$ the empty string, %
and $\delta_{x,y}=\delta_{xy}=1$ if $x=y$ and $0$ else is the Kronecker symbol. %
I generally omit separating commas if no confusion arises, in
particular in indices.
For any $x$ of suitable type (string,vector,set), I define
string $\v x = x_{1:l} = x_1...x_l$, %
sum $x_\p=\sum_j x_j$, union $x_*=\bigcup_j x_j$, and vector $\v x_\vi=(x_1,...,x_l)$, %
where $j$ ranges over the full range $\{1,...,l\}$
and $l=|{\rm x}|$ is the length or dimension or size of $\rm x$. %
$\hat x$ denotes an estimate of $x$. %
$\P(\cdot)$ denotes a probability over states and rewards or parts
thereof. I do not distinguish between random variables $X$ and
realizations $x$, and abbreviation $\P(x):=\P[X=x]$ never
leads to confusion.
%
More specifically, $m\in\SetN$ denotes the number of states, %
$i\in\{1,...,m\}$ any state index, %
$n\in\SetN$ the current time, %
and $t\in\{1,...,n\}$ any time in history. %
Further, in order not to get distracted at several places I gloss
over initial conditions or special cases where inessential. Also
0$*$undefined=0$*$infinity:=0.

\section{Feature Markov Decision Process ($\mathbf\Phi$MDP)}\label{secPhiMDP}

This section describes our formal setup. It consists of the
agent-environment framework and maps $\Phi$ from
observation-reward-action histories to MDP states. I call this
arrangement ``Feature MDP'' or short $\Phi$MDP.

\paradot{Agent-environment setup}
I consider the standard agent-environment setup \cite{Russell:03}
in which an {\em Agent} interacts with an {\em Environment}
The agent can choose from actions $a\in\A$ (e.g.\ limb movements)
and the environment provides (regular) observations $o\in\O$ (e.g.\
camera images) and real-valued rewards $r\in\R\subseteq\SetR$ to the
agent. The reward may be very scarce, e.g.\ just $+1$ ($-1$) for winning
(losing) a chess game, and 0 at all other times
\cite[Sec.6.3]{Hutter:04uaibook}.
This happens in cycles $t=1,2,3,...$: At time $t$, after observing
$o_t$ and receiving reward $r_t$, the agent takes action $a_t$ based on
history $h_t:=o_1 r_1 a_1...o_{t-1} r_{t-1} a_{t-1} o_t r_t$.
Then the next cycle $t+1$ starts. The agent's objective is to
maximize his long-term reward.
Without much loss of generality, I assume that $\R$ is finite.
Finiteness of $\R$ is lifted in
\cite{Hutter:09phidbn,Hutter:09phidbnx}. I also assume that $\A$ is
finite and small, which is restrictive. Part II deals with large
state spaces, and large (structured) action spaces can be dealt with in
a similar way. No assumptions are made on $\O$; it may be huge or
even infinite. Indeed, $\Phi$MDP has been specifically designed to
cope with huge observation spaces, e.g.\ camera images, which are
mapped to a small space of relevant states.

The agent and environment may be viewed as a pair or triple of interlocking
functions of the history $\H:=(\O\times\A\times\R)^*\times\O\times\R$:
\bqan
  & & \nq\Env:\H\times\A\leadsto\O\times\R, \quad   o_n r_n = \Env(h_{n-1}a_{n-1}),\hspace{25ex} \\
  & & \nq\!\!\!\!\!\Agent:\H\leadsto\A, \qquad\qquad\qquad a_n = \Agent(h_n),
\eqan
\begin{flushright}
\unitlength=1.2ex
\linethickness{0.4pt}
\begin{picture}(18,0)(0,-2) 
\thicklines\small
\put(3,3){\oval(6,2)[cc]\makebox(0,0)[cc]{Agent}}
\put(15,3){\oval(6,2)[cc]\makebox(0,0)[cc]{Env()}}
\put(3,2){\line(0,-1){2}}
\put(3,0){\line(1,0){12}}\put(9,0.5){\makebox(0,0)[cb]{$\boldsymbol a$\it\hspace{-1pt}ction}}
\put(15,0){\vector(0,1){2}}
\put(14,4){\line(0,1){2}}
\put(14,6){\line(-1,0){10}}\put(9,6){\makebox(0,0)[ct]{$\boldsymbol r$\it\hspace{-1.5pt}eward}}
\put(4,6){\vector(0,-1){2}}
\put(16,4){\line(0,1){4}}
\put(16,8){\line(-1,0){14}}\put(9,8){\makebox(0,0)[ct]{$\boldsymbol o$\it\hspace{-0.5pt}bservation}}
\put(2,8){\vector(0,-1){4}}
\end{picture}
\end{flushright}\vspace{-5ex}
where $\leadsto$ indicates that mappings $\to$ might be stochastic.

The goal of AI is to design agents that achieve high (expected)
reward over the agent's lifetime.

\paradot{(Un)known environments}
For known \Env(), finding the reward maximizing agent is a
well-defined and formally solvable problem
\cite[Chp.4]{Hutter:04uaibook}, with computational efficiency being
the ``only'' matter of concern. For most real-world AI problems
\Env() is at best partially known. For unknown \Env(), the meaning
of expected reward maximizing is even conceptually a challenge
\cite[Chp.5]{Hutter:04uaibook}.

Narrow AI considers the case where function \Env() is either known
(like planning in blocks world), or essentially known (like in chess, where
one can safely model the opponent as a perfect minimax player), or
\Env() belongs to a relatively small class of environments (e.g.\
elevator or traffic control).

The goal of AGI is to design agents that perform well in a large
range of environments \cite{Hutter:07iorx}, i.e.\ achieve high
reward over their lifetime with as little as possible assumptions
about Env(). A minimal necessary assumption is that the environment
possesses {\em some} structure or pattern \cite{Wolpert:97}.

From real-life experience (and from the examples below) we know that
usually we do not need to know the complete history of events in
order to determine (sufficiently well) what will happen next and to
be able to perform well. Let $\Phi(h)$ be such a ``useful'' summary of
history $h$.

\paradot{Generality of $\Phi$MDP}
The following examples show that many problems can be reduced
(approximately) to finite MDPs, thus showing that $\Phi$MDP can
deal with a large variety of problems:
In full-information {\em games} (like chess) with a static opponent,
it is sufficient to know the current state of the game
(board configuration) to play well (the history plays no role),
hence $\Phi(h_t)=o_t$ is a sufficient summary (Markov condition).
Classical {\em physics} is essentially predictable from the position and
velocity of objects at a single time, or equivalently from the
locations at two consecutive times, hence $\Phi(h_t)=o_{t-1}o_t$ is
a sufficient summary (2nd order Markov).
For {\em i.i.d.\ processes} of unknown probability (e.g.\ clinical trials
$\simeq$ Bandits), the frequency of observations
$\Phi(h_n)=(\sum_{t=1}^n\delta_{o_t o})_{o\in\O}$
is a sufficient statistic.
In a {\em POMDP planning} problem, the so-called belief vector at
time $t$ can be written down explicitly as some function of the complete
history $h_t$ (by integrating out the hidden states). $\Phi(h_t)$
could be chosen as (a discretized version of) this belief vector, showing
that $\Phi$MDP generalizes POMDPs.
Obviously, the {\em identity} $\Phi(h)=h$ is always sufficient but not
very useful, since \Env() as a function of $\H$ is hard to
impossible to ``learn''.

This suggests to look for $\Phi$ with small codomain, which allow
to learn/estimate/approximate $\Env$ by $\widehat\Env$ such that
$o_t r_t\approx\widehat\Env(\Phi(h_{t-1}))$ for $t=1...n$.

\paradot{Example}
Consider a robot equipped with a camera, i.e.\ $o$ is a pixel image.
Computer vision algorithms usually extract a set of features from
$o_{t-1}$ (or $h_{t-1}$), from low-level patterns to high-level
objects with their spatial relation. Neither is it possible nor
necessary to make a precise prediction of $o_t$ from summary
$\Phi(h_{t-1})$. An approximate prediction must and will do.
The difficulty is that the similarity measure ``$\approx$'' needs to
be context dependent. Minor image nuances are irrelevant when
driving a car, but when buying a painting it makes a huge difference
in price whether it's an original or a copy. Essentially only a
bijection $\Phi$ would be able to extract {\em all potentially}
interesting features, but such a $\Phi$ defeats its original
purpose.

\paradot{From histories to states}
It is of utmost importance to properly formalize the meaning of
``$\approx$'' in a general, domain-independent way.
Let $s_t:=\Phi(h_t)$ summarize all relevant information in history
$h_t$. I call $s$ a state or feature (vector) of $h$. ``Relevant''
means that the future is predictable from $s_t$ (and $a_t$) alone,
and that the relevant future is coded in $s_{t+1}s_{t+2}...$. So we
pass from the complete (and known) history $o_1 r_1 a_1...o_n r_n
a_n$ to a ``compressed'' history $sra_{1:n}\equiv s_1 r_1 a_1...s_n
r_n a_n$ and seek $\Phi$ such that $s_{t+1}$ is (approximately a
stochastic) function of $s_t$ (and $a_t$).
Since the goal of the agent is to maximize his rewards, the rewards
$r_t$ are always relevant, so they (have to) stay untouched (this
will become clearer below).

\paradot{The $\Phi$MDP}
The structure derived above is a classical Markov Decision Process
(MDP), but the primary question I ask is not the usual one of
finding the value function or best action or comparing different
models of a given state sequence. I ask how well can the
state-action-reward sequence generated by $\Phi$ be modeled as an
MDP compared to other sequences resulting from different $\Phi$.
A good $\Phi$ leads to a good model for predicting future rewards,
which can be used to find good actions that maximize the agent's
expected long-term reward.

\section{$\mathbf\Phi$MDP Coding and Evaluation}\label{secCE}

I first review a few standard codes and model selection methods for
i.i.d.\ sequences, subsequently adapt them to our situation, and
show that they are suitable in our context. I state my Cost function
for $\Phi$, and the $\Phi$ selection principle, and compare it to
the Minimum Description Length (MDL) philosophy.

\paradot{I.i.d.\ processes}
Consider i.i.d.\ $x_1...x_n\in\X^n$ for finite $\X=\{1,...,m\}$.
For known $\t_i=\P[x_t=i]$ we have
$\P(x_{1:n}|\v\t)=\t_{x_1}\cdot...\cdot\t_{x_n}$.
It is well-known that there exists a code (e.g.\ arithmetic or
Shannon-Fano) for $x_{1:n}$ of length $-\log\P(x_{1:n}|\v\t)$, which is
asymptotically optimal with probability one \cite[Thm.3.1]{Barron:85}.
This also easily follows from \cite[Thm.5.10.1]{Cover:06}. 

{\em MDL/MML code \cite{Gruenwald:07book,Wallace:05}:}
For unknown $\v\t$ we may use a frequency estimate $\hat\t_i=n_i/n$,
where $n_i=|\{t\leq n:x_t=i\}|$. Then it is easy to see that
$-\log\P(x_{1:n}|\v{\hat\t})=n\,H(\v{\hat\t})$, where
\beqn
  H(\v{\hat\t}) \;:=\; -\sum_{i=1}^{m}\hat\t_i\log\hat\t_i
  \qmbox{is the entropy of} \v{\hat\t}
\eeqn
($0\log 0:=0=:0\log{0\over 0}$). We also need to code $\v{\hat\t}$,
or equivalently $(n_i)$, which naively needs $\log n$ bits for each
$i$.
In general, a sample size of $n$ allows estimating parameters
only to accuracy $O(1/\sqrt{n})$, which is essentially equivalent to
the fact that $\log\P(x_{1:n}|\v{\hat\t}\pm
O(1/\sqrt{n}))-\log\P(x_{1:n}|\v{\hat\t}) = O(1)$.
This shows that it is sufficient to code each $\hat\t_i$ to accuracy
$O(1/\sqrt{n})$, which requires only $\fr12\log n+O(1)$ bits each.
Hence, given $n$ and ignoring $O(1)$ terms, the overall code length
(CL) of $x_{1:n}$ for unknown frequencies is
\beq\label{iidCodeL}
  \CL(x_{1:n}) \;\equiv\; \CLN(\v n) \;:=\;
  n\,H(\v n/n) + \fr{m-1}2\log n
  \qmbox{for} n>0 \qmbox{and} 0 \qmbox{else,}
\eeq
where $\v n=(n_1,...,n_m)$ and $n=n_+=n_1+...+n_m$.
We have assumed that $n$ is given, hence only $m-1$ of the $n_i$
need to be coded, since the $m$th one can be reconstructed from them
and $n$. The above is an exact code of $x_{1:n}$, which is optimal
(within $+O(1)$) for all i.i.d.\ sources.
This code may further be optimized by only coding $\hat\t_i$ for the
$m'=|\{i:n_i>0\}|\leq m$ non-empty categories, resulting in a code
of length
\beq\label{iidCodeLp}
  \CL'(\v n) \;:=\; n\,H(\v n/n) + \fr{m'-1}2\log n + m,
\eeq
where the $m$ bits
are needed to indicate which of the $\hat\t_i$ are coded. We refer to this
improvement as {\em sparse} code.

{\em Combinatorial code \cite{Li:08}:} A second way to code the data
is to code $\v n$ exactly, and then, since there are
$n!/n_1!...n_{m}!$ sequences $x_{1:n}$ with counts $\v n$, we can
easily construct a code of length $\log(n!/n_1!...n_{m}!)$ given $\v
n$ by enumeration, i.e.
\beqn
  \CL''(\v n) \;:=\; \log(n!/n_1!...n_{m}!) + (m\!-\!1)\log n
\eeqn
Within $\pm O(1)$ this code length also coincides with \req{iidCodeL}.

{\em Incremental code \cite{Willems:97}:} A third way is to use a
sequential estimate $\hat\t_i^{t+1}={t_i+\a\over t+m\a}$ based on
{\em known} past counts $t_i=|\{t'\leq t:x_{t'}=i\}|$, where $\a>0$
is some regularizer. Then
\beq\label{iidGamma}
 \P(x_{1:n}) \;=\; \hat\t_{x_1}^1\!\cdot...\cdot\hat\t_{x_n}^n
 \;=\; C_\alpha{\prod_{i=1}^{m}\Gamma(n_i+\a)\over\Gamma(n+m\a)},
 \qquad C_\alpha:={\Gamma(m\a)\over\Gamma(\a)^{m}}
\eeq
where $\Gamma$ is the Gamma function. The logarithm of this
expression again essentially reduces to \req{iidCodeL} (for any
$\a>0$, typically $\fr12$ or 1), which can also be written as
\beqn
  \CL'''(\v n)=\sum_{i:n_i>0}\ln\Gamma(n_i)-\ln\Gamma(n)+O(1)
  \qmbox{if} n>0 \qmbox{and 0 else.}
\eeqn

{\em Bayesian code \cite{Schwarz:78,MacKay:03}:} A fourth (the
Bayesian) way is to assume a Dirichlet($\a$) prior over $\v\t$. The
marginal distribution (evidence) is identical to \req{iidGamma} and
the Bayesian Information Criterion (BIC) approximation leads to code
\req{iidCodeL}.

{\em Conclusion:} All four methods lead to essentially the same code
length. The references above contain rigorous derivations.
In the following I will ignore the $O(1)$ terms and refer to
\req{iidCodeL} simply as {\em the} code length. Note that $x_{1:n}$
is coded exactly (lossless). Similarly (see MDP below) sampling
models more complex than i.i.d.\ may be considered, and the one that
leads to the shortest code is selected as the best model
\cite{Gruenwald:07book}.

\paradot{MDP definitions}
Recall that a sequence $sra_{1:n}$ is said to be sampled from an MDP
$(\S,\A,T,R)$ iff the probability of $s_t$ only depends on $s_{t-1}$
and $a_{t-1}$; and $r_t$ only on
$s_{t-1}$, $a_{t-1}$, and $s_t$. That is,
\bqan
  \P(s_t|h_{t-1}a_{t-1}) \;=\; \P(s_t|s_{t-1},a_{t-1}) &=:& T_{s_{t-1}s_t}^{a_{t-1}} \\
         \P(r_t|h_t) \;=\; \P(r_t|s_{t-1},a_{t-1},s_t) &=:& R_{s_{t-1}s_t}^{a_{t-1}r_t}
\eqan
In our case, we can identify the state-space $\S$ with the states
$s_1,...,s_n$ ``observed'' so far. Hence $\S=\{s^1,...,s^m\}$ is
finite and typically $m\ll n$, since states repeat. Let
$s\stackrel{a}\to s'(r')$ be shorthand for ``action $a$ in state $s$
resulted in state $s'$ (reward $r'$)''. Let $\T_{ss'}^{ar'}:=\{t\leq
n:s_{t-1}=s, a_{t-1}=a, s_t=s', r_t=r'\}$ be the set of times $t-1$
at which $s\stackrel{a}\to s'r'$, and
$n_{ss'}^{ar'}:=|\T_{ss'}^{ar'}|$ their number
($n_{\p\p}^{\p\p}=n$).

\paradot{Coding MDP sequences}
For some fixed $s$ and $a$, consider the subsequence
$s_{t_1}...s_{t_{n'}}$ of states reached from $s$ via $a$
($s\stackrel{a}\to s_{t_i}$), i.e.\
$\{t_1,...,t_{n'}\}=\T_{s*}^{a*}$, where $n'=n_{s\p}^{a\p}$. By
definition of an MDP, this sequence is i.i.d.\ with $s'$ occurring
$n'_{s'}:=n_{ss'}^{a\p}$ times. By \req{iidCodeL} we can code this
sequence in $\CL(\v n')$ bits. The whole sequence $s_{1:n}$ consists
of $|\S\times\A|$ i.i.d.\ sequences, one for each
$(s,a)\in\S\times\A$. We can join their codes and get a total code
length
\beq\label{sCode}
  \CL(s_{1:n}|a_{1:n}) \;=\; \sum_{s,a} \CLN(\v n_{s\vi}^{a\p})
\eeq
If instead of \req{iidCodeL} we use the improved sparse code \req{iidCodeLp},
non-occurring transitions $s\stackrel{a}\to s'r'$ will contribute
only one bit rather than $\fr12\log n$ bits to the code, so that
large but sparse MDPs get penalized less.

Similarly to the states we code the rewards. There are different
``standard'' reward models. I consider only the simplest case of a
small discrete reward set $\R$ like $\{0,1\}$ or $\{-1,0,+1\}$ here
and defer generalizations to $\SetR$ and a discussion of variants to
the $\Phi$DBN model \cite{Hutter:09phidbn}.
By the MDP assumption, for each $(s,a,s')$ triple, the rewards at times
$\T_{ss'}^{a*}$ are i.i.d. Hence they can be coded in
\beq\label{rCode}
  \CL(r_{1:n}|s_{1:n},a_{1:n}) \;=\; \sum_{s,a,s'} \CLN(\v n_{ss'}^{a\vi})
\eeq
bits.
In order to increase the statistics it might be better to
treat $r_t$ as a function of $s_t$ only. This is not restrictive,
since dependence on $s_{t-1}$ and $a_{t-1}$ can be mimicked by
coding aspects into an enlarged state space.

\paradot{Reward$\mathbf\leftrightarrow$state trade-off}
Note that the code for $\v r$ depends on $\v s$. Indeed we may
interpret the construction as follows: Ultimately we/the agent cares
about the reward, so we want to measure how well we can predict the
rewards, which we do with \req{rCode}.
But this code depends on $\v s$, so we need a code for $\v s$ too,
which is \req{sCode}. To see that we need both parts consider two
extremes.

A simplistic state transition model (small $|\S|$) results in a
short code for $\v s$. For instance, for $|\S|=1$, nothing needs to be
coded and \req{sCode} is identically zero. But this obscures
potential structure in the reward sequence, leading to a long code
for $\v r$.

On the other hand, the more detailed the state transition model
(large $|\S|$) the easier it is to predict and hence compress $\v
r$. But a large model is hard to learn, i.e.\ the code for $\v s$
will be large. For instance for $\Phi(h)=h$, no state repeats and
the frequency-based coding breaks down.

\paradot{$\Phi$ selection principle}
Let us define the {\em Cost} of $\Phi:\H\to\S$ on $h_n$ as the
length of the $\Phi$MDP code for $\v s\v r$ given $\v a$ plus a
complexity penalty $\CL(\Phi)$ for $\Phi$:
\bqa\label{costphi}
  & & \nq\nq\Cost(\Phi|h_n) \;:=\;
      \CL(s_{1:n}|a_{1:n}) + \CL(r_{1:n}|s_{1:n},a_{1:n}) + \CL(\Phi),
\\ \nonumber
  & & \nq\nq\qmbox{where} s_t=\Phi(h_t)\qmbox{and} h_t=ora_{1:t-1}o_t r_t
\eqa
The discussion above suggests that the minimum of the joint code
length \req{sCode} and \req{rCode} is attained for a $\Phi$ that
keeps all and only relevant information for predicting rewards. Such
a $\Phi$ may be regarded as best explaining the rewards.
I added an additional complexity penalty $\CL(\Phi)$ for $\Phi$ such
that from the set of $\Phi$ that minimize \req{sCode}+\req{rCode}
(e.g.\ $\Phi$'s identical on $(\O\times\R\times\A)^n$ but different on
longer histories) the simplest one is selected. The penalty is
usually some code-length or log-index of $\Phi$. This conforms with
Ockham's razor and the MDL philosophy.
So we are looking for a $\Phi$ of minimal cost:
\beq\label{bestphi}
   \Phi^{best} \;:=\; \arg\min_\Phi\{ \Cost(\Phi|h_n) \}
\eeq
If the minimization is restricted to some small class of reasonably
simple $\Phi$, $\CL(\Phi)$ in \req{costphi} may be dropped. The
state sequence generated by $\Phi^{best}$ (or approximations
thereof) will usually only be approximately MDP. While
$\Cost(\Phi|h)$ is an optimal code only for MDP sequences, it still
yields good codes for approximate MDP sequences. Indeed,
$\Phi^{best}$ balances closeness to MDP with simplicity. The primary
purpose of the simplicity bias is {\em not} computational
tractability, but generalization ability
\cite{Legg:08,Hutter:04uaibook}.

\paradot{Relation to MDL et al}
In unsupervised learning (clustering and density estimation) and
supervised learning (regression and classification), penalized
maximum likelihood criteria \cite[Chp.7]{Hastie:01} like BIC
\cite{Schwarz:78}, MDL \cite{Gruenwald:07book}, and MML
\cite{Wallace:05} have successfully been used for semi-parametric
model selection. It is far from obvious how to apply them in RL.
Indeed, our derived Cost function cannot be interpreted as a usual
model+data code length. The problem is the following:

Ultimately we do not care about the observations but the rewards.
The rewards depend on the states, but the states are arbitrary in
the sense that they are model-dependent functions of the bare data
(observations). The existence of these unobserved states is what
complicates matters, but their introduction is necessary in order to
model the rewards.
For instance, $\Phi$ is actually not needed for coding $\v r\v s|\v
a$, so from a strict coding/MDL perspective, $\CL(\Phi)$ in
\req{costphi} is redundant. Since $\v s$ is some ``arbitrary''
construct of $\Phi$, it is better to regard \req{costphi} as a code
of $\v r$ only. Since the agent chooses his actions, $\v a$ need not
be coded, and $\v o$ is not coded, because they are only of indirect
importance.

The Cost() criterion is strongly motivated by the rigorous MDL
principle, but invoked outside the usual
induction/modeling/prediction context.

\section{A Tiny Example}\label{secTE}

The purpose of the tiny example in this section is to provide enough
insight into how and why $\Phi$MDP works to convince the reader that
our $\Phi$ selection principle is reasonable.

\paradot{Example setup}
I assume a simplified MDP model in which reward $r_t$ only depends
on $s_t$, i.e.\
\beq\label{rCodeEx}
  \CL(r_{1:n}|s_{1:n},a_{1:n}) \;=\; \sum_{s'} \CLN(\v n_{\p s'}^{\p\vi})
\eeq
This allows us to illustrate $\Phi$MDP on a tiny example. The
same insight is gained using \req{rCode} if an analogous larger
example is considered. Furthermore I set $\CL(\Phi)\equiv 0$.

Consider binary observation space $\O=\{0,1\}$, quaternary reward
space $\R=\{0,1,2,3\}$, and a single action $\A=\{0\}$. Observations
$o_t$ are independent fair coin flips, i.e.\ Bernoulli($\fr12$), and
reward $r_t=2o_{t-1}+o_t$ a deterministic function of the two most
recent observations.

\paradot{Considered features}
As features $\Phi$ I consider $\Phi_k:\H\to\O^k$ with
$\Phi_k(h_t)=o_{t-k+1}...o_t$ for various $k=0,1,2,...$ which regard
the last $k$ observations as ``relevant''.
Intuitively $\Phi_2$ is the best observation summary, which I confirm
below. The state space $\S=\{0,1\}^k$ (for sufficiently large $n$).
The $\Phi$MDPs for $k=0,1,2$ are as follows.

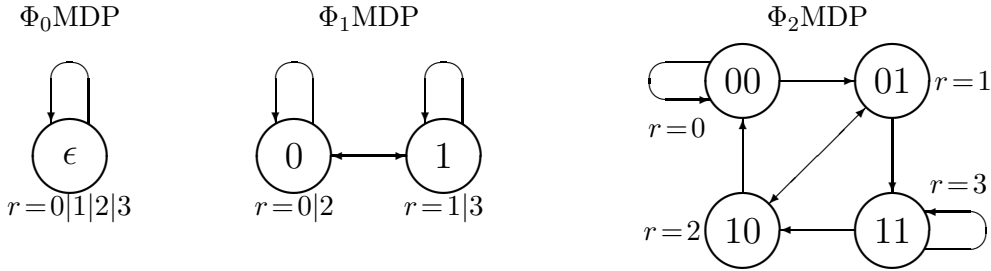
\begin{center}
\small\unitlength=3ex
\linethickness{0.4pt}
\begin{picture}(27,7)(0,0.3)
\thinlines
\put(2,7){\makebox(0,0)[ct]{$\Phi_0$MDP}}
\thicklines\put(2,3){\circle{2}}\thinlines
\put(2,3){\makebox(0,0)[cc]{\large$\epstr$}}
\put(2,2){\makebox(0,0)[ct]{$r=0|1|2|3$}}
\put(2,5){\oval(1,1)[t]}
\put(1.5,5){\vector(0,-1){1.13}}
\put(2.5,5){\line(0,-1){1.13}}

\put(10,7){\makebox(0,0)[ct]{$\Phi_1$MDP}}
\thicklines\put(8,3){\circle{2}}\thinlines
\put(8,3){\makebox(0,0)[cc]{\large 0}}
\put(8,2){\makebox(0,0)[ct]{$r=0|2$}}
\put(8,5){\oval(1,1)[t]}
\put(7.5,5){\vector(0,-1){1.13}}
\put(8.5,5){\line(0,-1){1.13}}
\put(10,3){\vector(1,0){1}}
\put(10,3){\vector(-1,0){1}}
\thicklines\put(12,3){\circle{2}}\thinlines
\put(12,3){\makebox(0,0)[cc]{\large 1}}
\put(12,2){\makebox(0,0)[ct]{$r=1|3$}}
\put(12,5){\oval(1,1)[t]}
\put(11.5,5){\vector(0,-1){1.13}}
\put(12.5,5){\line(0,-1){1.13}}

\put(22,7){\makebox(0,0)[ct]{$\Phi_2$MDP}}
\thicklines\put(20,5){\circle{2}}\thinlines
\put(20,5){\makebox(0,0)[cc]{\large 00}}
\put(19,4){\makebox(0,0)[rt]{$r=0$}}
\put(18,5){\oval(1,1)[l]}
\put(18,4.5){\vector(1,0){1.13}}
\put(18,5.5){\line(1,0){1.13}}
\thicklines\put(24,1){\circle{2}}\thinlines
\put(24,1){\makebox(0,0)[cc]{\large 11}}
\put(25,2){\makebox(0,0)[lb]{$r=3$}}
\put(26,1){\oval(1,1)[r]}
\put(26,1.5){\vector(-1,0){1.13}}
\put(26,0.5){\line(-1,0){1.13}}
\thicklines\put(24,5){\circle{2}}\thinlines
\put(24,5){\makebox(0,0)[cc]{\large 01}}
\put(25,5){\makebox(0,0)[lc]{$\,r=1$}}
\thicklines\put(20,1){\circle{2}}\thinlines
\put(20,1){\makebox(0,0)[cc]{\large 10}}
\put(19,1){\makebox(0,0)[rc]{$\,r=2\,$}}
\put(21,5){\vector(1,0){2}}
\put(24,4){\vector(0,-1){2}}
\put(23,1){\vector(-1,0){2}}
\put(20,2){\vector(0,1){2}}
\put(22,3){\vector(1,1){1.3}}
\put(22,3){\vector(-1,-1){1.3}}
\end{picture}
\end{center}

\paranodot{$\mathbf\Phi_2$MDP}
with all non-zero transition probabilities being 50\% is an exact
representation of our data source. The missing arrow (directions)
are due to the fact that $s=o_{t-1}o_t$ can only lead to $s'=o'_t
o'_{t+1}$ for which $o'_t=o_t$, denoted by $s*=*s'$ in the
following. Note that $\Phi$MDP does not ``know'' this and has to
learn the (non)zero transition probabilities. Each state has two
successor states with equal probability, hence generates (see
previous paragraph) a Bernoulli($\fr12$) state subsequence and a
constant reward sequence, since the reward can be computed from the
state = last two observations. Asymptotically, all four states occur
equally often, hence the sequences have approximately the same
length $n/4$.

In general, if $\v s$ (and similarly $\v r$) consists of
$x\in\SetN$ i.i.d.\ subsequences of equal length $n/x$ over $y\in\SetN$
symbols, the code length \req{sCode} (and similarly \req{rCodeEx}) is
\bqan
  \CL(\v s|\v a;x_y)   &=& \textstyle n\log y + x{|\S|-1\over 2}\log{n\over x} \\
  \CL(\v r|\v s,\v a;x_y) &=& \textstyle n\log y + x{|\R|-1\over 2}\log{n\over x}
\eqan
where the extra argument $x_y$ just indicates the sequence property.
So for $\Phi_2$MDP we get
\beqn
  \CL(\v s|\v a;4_2) = n+6\log\fr n4 \qmbox{and}
  \CL(\v r|\v s,\v a;4_1) = 6\log\fr n4
\eeqn
The log-terms reflect the required memory to code the MDP structure
and probabilities. Since each state has only 2 realized/possible
successors, we need $n$ bits to code the state sequence. The reward
is a deterministic function of the state, hence needs no memory to
code given $\v s$.

\paranodot{The $\mathbf\Phi_0$MDP}
throws away all observations (left figure above), hence
$\CL(\v s|\v a;1_1)=0$. While the reward sequence is {\em not} i.i.d.\
(e.g.\ $r_{t+1}=3$ cannot follow $r_t=0$), $\Phi_0$MDP has no choice
regarding them as i.i.d., resulting in $\CL(\v s|\v a;1_4)=2n+\fr32\log
n$.

\paranodot{The $\mathbf\Phi_1$MDP}
model is an interesting compromise (middle figure above). The state
allows a partial prediction of the reward: State 0 allows rewards 0
and 2; state 1 allows rewards 1 and 3. Each of the two states
creates a Bernoulli($\fr12$) state successor subsequence and a
binary reward sequence, wrongly presumed to be Bernoulli($\fr12$).
Hence $\CL(\v s|\v a;2_2)=n+\log\fr n2$ and $\CL(\v r|\v s,\v
a;2_2)=n+3\log\fr n2$.

\paradot{Summary}
The following table summarizes the results for general $k=0,1,2$ and beyond:
\beqn\small\arraycolsep1pt\textmuskip
\begin{array}{c||c|c|c|c|c|c|c|c|c}
 k & \S                & |\S| & n_{ss'}^{0\p} & n_{\p s'}^{\p r'} & n_{s\p}^{0\p}\!\!=\!n_{\p s'}^{\p\p} & \v s+\v r & \CL(\v s|\v a) & \CL(\v r|\v s,\v a) & \Cost(\Phi|h) \\ \hline\hline
 0 & \{\epstr\}        &   1    & n                       & n/4                          & n     & 1_1+1_4           & 0             & 2n\!+\!\fr32\log n & 2n\!+\!\fr32\log n \\
 1 & \{0,1\}           &   2    & n/4                     & \fr n4\delta_{r'\!-\!s'=0|1} & n/2   & 2_2+2_2           & n+\log\fr n2  & n+3\log\fr n2  & 2n+4\log\fr n2 \\
 2 & \{{00,01\atop 10,11}\} & 4 & \fr n8\delta_{s*,*s'}   & \fr n4\delta_{r'\widehat=s'} & n/4   & 4_2+4_1           & n+6\log\fr n4 & 6\log\fr n4    & n+12\log\fr n4 \\
^{_\geq}\!2 & \{0,1\}^k & 2^k & \fr{n\delta_{s*,*s'}}{2^{k+1}} & \fr n4\delta_{r'\widehat=s'} & n/2^k & 2^k_2+2^k_2 & n+{2^k\!-\!1\over 2^{1\!-k}}\log\!\fr{n}{2^k} & \fr32 2^k\log\fr{n}{s^k} & n+{2^k\!+\!2\over 2^{1\!-k}}\log\!\fr{n}{2^k}
\end{array}
\eeqn
The notation of the $\v s+\v r$ column follows the one used above in
the text ($x_y$ for $\v s$ and $\v r$). $r'\widehat=s'$ means that
$r'$ is the correct reward for state $s'$. The last column is the
sum of the two preceding columns. The part linear in $n$ is the code
length for the state/reward sequence. The part logarithmic in $n$ is
the code length for the transition/reward probabilities of the MDP;
each parameter needs $\fr12\log n$ bits.
For large $n$, $\Phi_2$ results in the shortest code, as
anticipated. The ``approximate'' model $\Phi_1$ is just not good
enough to beat the vacuous model $\Phi_0$, but in more realistic
examples some approximate model usually has the shortest code. In
\cite{Hutter:09phidbn} I show on a more complex example how
$\Phi^{best}$ will store long-term information in a POMDP
environment.

\section{\Cost($\mathbf\Phi$) Minimization}\label{secCM}

So far I have reduced the reinforcement learning problem to a formal
$\Phi$-optimization problem. This section briefly explains what we
have gained by this reduction, and provide some general information
about problem representations, stochastic search, and $\Phi$
neighborhoods. Finally I present a simplistic but concrete algorithm
for searching context tree MDPs.

\paradot{$\mathbf\Phi$ search}
I now discuss how to find good summaries $\Phi$. The introduced
generic cost function $\Cost(\Phi|h_n)$, based on only the known
history $h_n$, makes this a well-defined task that is completely
decoupled from the complex (ill-defined) reinforcement learning
objective. This reduction should not be under-estimated. We can
employ a wide range of optimizers and do not even have to worry
about overfitting. The most challenging task is to come up with
creative algorithms proposing $\Phi$'s.

There are many optimization methods: Most of them are search-based:
random, blind, informed, adaptive, local, global, population based,
exhaustive, heuristic, and other search methods \cite{Aarts:97}.
Most are or can be adapted to the structure of the objective
function, here $\Cost(\cdot|h_n)$. Some exploit the structure more
directly (e.g.\ gradient methods for convex functions). Only in very
simple cases can the minimum be found analytically (without search).

Most search algorithms require the specification of a neighborhood
relation or distance between candidate $\Phi$, which I define in
the 2nd next paragraph.

\paranodot{Problem representation}
can be important: Since $\Phi$ is a discrete function, searching
through (a large subset of) all computable functions, is a
non-restrictive approach. Variants of Levin search
\cite{Schmidhuber:04oops,Hutter:04uaibook} and genetic programming
\cite{Koza:92,Banzhaff:98} and recurrent neural networks
\cite{Pearlmutter:89,Raedt:08} are the major approaches in this
direction.

A different representation is as follows: $\Phi$ effectively
partitions the history space $\H$ and identifies each partition with
a state. Conversely any partition of $\H$ can (up to a renaming of
states) uniquely be characterized by a function $\Phi$.
Formally, $\Phi$ induces a (finite) partition
$\bigcup_s\{h':\Phi(h')=s\}$ of $\H$, where
$s$ ranges over the codomain of $\Phi$. Conversely, any partition of
$\H={\cal B}_1\dot\cup...\dot\cup{\cal B}_m$
induces a function $\Psi(h')=i$ iff $h'\in{\cal B}_i$, which is
equivalent to $\Phi$ apart from an irrelevant permutation of the
codomain (renaming of states).

State aggregation methods have been suggested earlier for solving
large-scale MDP planning problems by grouping (partitioning) similar
states together, resulting in (much) smaller block MDPs
\cite{Givan:03}. But the used bi-simulation metrics
require knowledge of the MDP transition probabilities, while
our Cost criterion does not.

Decision trees/lists/grids/etc.\ are essentially space partitioners.
The most powerful versions are rule-based, in which logical
expressions recursively divide domain $\H$ into
``true/false'' regions \cite{Dzeroski:01,Sanner:09}.

\paradot{$\Phi$ neighborhood relation}
A natural ``minimal'' change of a partition is to subdivide=split a
partition or merge (two) partitions. Moving elements from one
partition to another can be implemented as a split and merge
operation. In our case this corresponds to splitting and merging
states (state refinement and coarsening).
Let $\Phi'$ split some state $s^a\in\S$ of $\Phi$ into
$s^b,s^c\not\in\S$
\beqn
  \Phi'(h) \;:=\; \left\{ {\Phi(h) \quad\qmbox{if} \Phi(h)\neq s^a \atop
                           s^b \mbox{ or } s^c \qmbox{if} \Phi(h)=s^a} \right.
\eeqn
where the histories mapped to state $s^a$ are distributed among $s^b$ and
$s^c$ according to some splitting rule (e.g.\ randomly). The new
state space is $\S'=\S\setminus\{s^a\}\cup\{s^b,s^c\}$.
Similarly $\Phi'$ merges states $s^b,s^c\in\S$ into $s^a\not\in\S$ if
\beqn
  \Phi'(h) \;:=\; \left\{ {\Phi(h) \qmbox{if} \Phi(h)\neq s^a\qquad \atop
                           s^a \quad\;\qmbox{if} \Phi(h)=s^b \mbox{ or } s^c} \right.
\eeqn
where $\S'=\S\setminus\{s^b,s^c\}\cup\{s^s\}$. We can regard
$\Phi'$ as being a neighbor of or similar to $\Phi$.

\paradot{Stochastic $\Phi$ search}
Stochastic search is the method of choice for high-dimensional
unstructured problems. Monte Carlo methods can actually be highly
effective, despite their simplicity \cite{Liu:02,Fishman:03}. The
general idea is to randomly choose a neighbor $\Phi'$ of $\Phi$ and
replace $\Phi$ by $\Phi'$ if it is better, i.e.\ has smaller Cost.
Even if $\Cost(\Phi'|h)>\Cost(\Phi|h)$ we may keep $\Phi'$, but only
with some (in the cost difference exponentially) small probability.
Simulated annealing is a version which minimizes $\Cost(\Phi|h)$.
Apparently, $\Phi$ of small cost are (much) more likely to occur
than high cost $\Phi$.

\paradot{Context tree example}
The $\Phi_k$ in Section \ref{secTE} depended on the last $k$
observations. Let us generalize this to a context dependent variable
length:
Consider a finite complete suffix free set of strings (= prefix tree
of reversed strings) $\S\subset\O^*$ as our state space (e.g.\
$\S=\{0,01,011,111\}$ for binary $\O$), and define $\Phi_\S(h_n):=s$
iff $o_{n-|s|+1:n}=s\in\S$, i.e.\ $s$ is the part of
the history regarded as relevant.
State splitting and merging works as follows: For binary $\O$, if
history part $s\in\S$ of $h_n$ is deemed too short, we replace $s$
by $0s$ and $1s$ in $\S$, i.e.\ $\S'=\S\setminus\{s\}\cup\{0s,1s\}$.
If histories $1s,0s\in\S$ are deemed too long, we replace them by
$s$, i.e.\ $\S'=\S\setminus\{0s,1s\}\cup\{s\}$.
Large $\O$ might be coded binary and then treated similarly.
For small $\O$ we have the following simple $\Phi$-optimizer:

\pagebreak[3]
\def\algitsep{\itemsep=0ex}
\begin{list}{}{\parskip=0ex\parsep=0ex\algitsep\leftmargin=0ex\labelwidth=0ex}
  \item {\bf\boldmath $\Phi$Improve($\Phi_\S,h_n$)}
  \begin{list}{}{\parskip=0ex\parsep=0ex\algitsep\leftmargin=2ex\labelwidth=1ex\labelsep=1ex}
    \item[$\lceil$] Randomly choose a state $s\in\S$;
    \item Let $p$ and $q$ be uniform random numbers in $[0,1]$;
    \item if $(p>1/2)$ then split $s$ i.e.\ $S'=S\setminus\{s\}\cup\{os:o\in\O\}$
    \item else if $\{os':o\in\O\}\subseteq\S\quad$ ({\it $s'$ is $s$ without the first symbol})
    \item then merge them, i.e.\ $S'=S\setminus\{os':o\in\O\}\cup\{s'\}$;
    \item if $(\Cost(\Phi_\S|h_n)-\Cost(\Phi_{\S'}|h_n) > \log(q))$ then $\S:=\S'$;
    \item [$\lfloor$] {\bf\boldmath return ($\Phi_\S$); }
  \end{list}
\end{list}
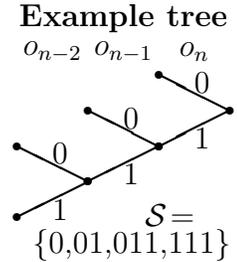
\begin{flushright}
\unitlength=1.3ex
\linethickness{0.4pt}
\begin{picture}(12,6)(0,-10) 
\thicklines
\put(6,14){\makebox(0,0)[ct]{\bf Example tree}}
\put(2,11.3){\makebox(0,0)[cc]{$o_{n-2}$}}
\put(6,11.3){\makebox(0,0)[cc]{$o_{n-1}$}}
\put(10,11.3){\makebox(0,0)[cc]{$o_n$}}
\put(10,1.5){\makebox(0,0)[rb]{$\S=$}}
\put(12,-0.5){\makebox(0,0)[rb]{$\{0,01,011,111\}$}}
\multiput(0,0)(4,2){3}{
  \put(4,4){\line(-2,1){4}}
  \put(4,4){\line(-2,-1){4}}
  \put(2,5){\makebox(0,0)[lb]{0}}
  \put(2,3){\makebox(0,0)[lt]{1}}
  \put(4,4){\circle*{0.4}}
  \put(0,6){\circle*{0.4}}
  \put(0,2){\circle*{0.4}}
} %
\end{picture}
\end{flushright}\vspace{-9ex}

The idea of using suffix trees as state space is from
\cite{McCallum:96} (see also \cite{Ring:94}).
It might be interesting to compare the local split/merge criterion
of \cite{McCallum:96} with our general global Cost criterion. On the
other hand, due to their limitation, suffix trees are currently out
of vogue.

\section{Exploration \& Exploitation}\label{secEE}

Having obtained a good estimate $\hat\Phi$ of $\Phi^{best}$ in the
previous section, we can/must now determine a good action for our
agent. For a finite MDP with known transition probabilities, finding
the optimal action is routine. For estimated probabilities we run
into the infamous exploration-exploitation problem, for which
promising approximate solutions have recently been suggested
\cite{Szita:08}. At the end of this section I present the overall
algorithm for our $\Phi$MDP agent.

\paradot{Optimal actions for known MDPs}
For a known finite MDP $(\S,\A,T,R,\gamma)$, the maximal achievable
(``optimal'') expected future discounted reward sum, called ($Q$)
$V\!$alue (of action $a$) in state $s$, satisfies the following
(Bellman) equations \cite{Sutton:98}
\beq\label{BellmanEq}
  Q_s^{*a} \;=\; \sum_{s'} T_{ss'}^a[R_{ss'}^a + \gamma V_{s'}^*]
  \qmbox{and} V_s^*=\max_a Q_s^{*a}
\eeq
where $0<\g<1$ is a discount parameter, typically close to 1. See
\cite[Sec.5.7]{Hutter:04uaibook} for proper choices. The equations
can be solved by a simple (e.g.\ value or policy) iteration process
or various other methods or in guaranteed polynomial time by dynamic
programming \cite{Puterman:94}.
The optimal next action is
\beq\label{BellmanSol}
  a_n := \arg\max_a Q_{s_n}^{*a}
\eeq

\paradot{Estimating the MDP}
We can estimate the transition probability $T$ by
\beq\label{hatT}
  \hat T_{ss'}^a \;:=\; {n_{ss'}^{a\p}\over n_{s\p}^{a\p}}
  \qmbox{if} n_{s\p}^{a\p}>0 \qmbox{and} 0 \qmbox{else.}
\eeq
It is easy to see that the Shannon-Fano code of $s_{1:n}$ based on
$\P_{\smash{\!\hat T}}(s_{1:n}|a_{1:n})=\prod_{t=1}^n\hat
T_{s_{t-1}s_t}^{a_{t-1}}$ plus the code of the (non-zero) transition
probabilities $\hat T_{ss'}^a$ to relevant accuracy
$O(1/\sqrt{n_{s\p}^{a\p}})$ has length \req{sCode}, i.e.\
the frequency estimate \req{hatT} is consistent with the
attributed code length.
The expected reward can be estimated as
\beq\label{hatR}
  \hat R_{ss'}^a := \sum_{r'\in\R}\hat R_{ss'}^{ar'} r',\qquad
  \hat R_{ss'}^{ar'} := {n_{ss'}^{ar'}\over n_{ss'}^{a\p}}
\eeq

\paradot{Exploration}
Simply replacing $T$ and $R$ in \req{BellmanEq} and \req{BellmanSol}
by their estimates \req{hatT} and \req{hatR} can lead
to very poor behavior, since parts of the state space may never be
explored, causing the estimates to stay poor.

Estimate $\hat T$ improves with increasing $n_{s\p}^{a\p}$, which
can (only) be ensured by trying all actions $a$ in all states $s$
sufficiently often. But the greedy policy above has no incentive to
explore, which may cause the agent to perform very poorly: The agent
stays with what he {\em believes} to be optimal without trying to
solidify his belief.
For instance, if treatment $A$ cured the first patient, and
treatment $B$ killed the second, the greedy agent will stick to treatment
$A$ and not explore the possibility that $B$ may just have failed
due to bad luck.
Trading off exploration versus exploitation
optimally is computationally intractable
\cite{Hutter:04uaibook,Poupart:06,Ross:08bayes} in all but extremely
simple cases (e.g.\ Bandits \cite{Berry:85,Kumar:86}).
Recently, polynomially optimal algorithms (Rmax,E3,OIM) have been
invented \cite{Kearns:98,Brafman:02,Szita:08}: An agent is more explorative if
he expects a high reward in the unexplored regions. We can
``deceive'' the agent to believe this by adding another
``absorbing'' high-reward state $s^e$ to $\S$, not in the range of
$\Phi(h)$, i.e.\ never observed. Henceforth, $\S$ denotes the
extended state space. For instance $+$ in \req{hatT} now includes
$s^e$. We set
\beq\label{extnR}
  n_{ss^e}^a=1,\quad n_{s^e s}^a=\delta_{s^e s},\quad R_{ss^e}^a=R_{max}^e
\eeq
for all $s,a$, where exploration bonus $R_{max}^e$ is polynomially (in
$(1-\g)^{-1}$ and $|\S\times\A|$) larger than $\max\R$
\cite{Szita:08}.

Now compute the agent's action by \req{BellmanEq}-\req{hatR} but for
the extended $\S$. The optimal policy $p^*$ tries to find a chain of
actions and states that likely leads to the high reward absorbing
state $s^e$. Transition $\hat T_{ss^e}^a=1/n_{s\p}^a$ is only ``large'' for
small $n_{s\p}^a$, hence $p^*$ has a bias towards unexplored
(state,action) regions. It can be shown that this algorithm makes
only a polynomial number of sub-optimal actions.

The overall algorithm for our $\Phi$MDP agent is as follows.

\pagebreak[3]\def\algitsep{\itemsep=0ex}
\begin{list}{}{\parskip=0ex\parsep=0ex\algitsep\leftmargin=0ex\labelwidth=0ex}
  \item {\bf\boldmath $\Phi$MDP-Agent($\A,\R$)}
  \begin{list}{}{\parskip=0ex\parsep=0ex\algitsep\leftmargin=2ex\labelwidth=1ex\labelsep=1ex}
    \item[$\lceil$] Initialize $\Phi\equiv\Phi'\equiv\epstr$; $\;\S=\{\epstr\}$; $\;h_0=a_0=r_0=\epstr$;
    \item for $n=1,2,3,...$
    \begin{list}{}{\parskip=0ex\parsep=0ex\algitsep\leftmargin=2ex\labelwidth=1ex\labelsep=1ex}
      \item[$\lceil$] Choose e.g.\ $\g=1-1/n$;
      \item Set $R_{max}^e=$Polynomial$((1-\g)^{-1},|\S\times\A|)\cdot\max\R$;
      \item While waiting for $o_n$ and $r_n$
      \begin{list}{}{\parskip=0ex\parsep=0ex\algitsep\leftmargin=2ex\labelwidth=1ex\labelsep=1ex}
        \item[$\lceil$] $\Phi':=\Phi$Improve($\Phi',h_{n-1}$);
        \item [$\lfloor$] If $\Cost(\Phi'|h_{n-1})<\Cost(\Phi|h_{n-1})$ then $\Phi:=\Phi'$;
      \end{list}
      \item Observe $o_n$ and $r_n$; $\;h_n:=h_{n-1}a_{n-1}r_{n-1}o_n r_n$;
      \item $s_n:=\Phi(h_n)$; $\;\S:=\S\cup\{s_n\}$;
      \item Compute action $a_n$ from Equations \req{BellmanEq}-\req{extnR};
    \end{list}
    \item [$\lfloor$] $\lfloor$ Output action $a_n$;
  \end{list}
\end{list}

\section{Improved Cost Function}\label{secICF}

As discussed, we ultimately only care about (modeling) the rewards,
but this endeavor required introducing and coding states. The
resulting Cost($\Phi|h$) function is a code length of not only the
rewards but also the ``spurious'' states. This likely leads to a too
strong penalty of models $\Phi$ with large state spaces $\S$. The
proper Bayesian formulation developed in this section allows to
``integrate'' out the states. This leads to a code for the rewards
only, which better trades off accuracy of the reward model and state
space size.

For an MDP with transition and reward probabilities $T_{ss'}^a$ and
$R_{ss'}^{ar'}$, the probabilities of the state and reward sequences
are
\beqn
  \P_{\!T}(s_{1:n}|a_{1:n}) = \prod_{t=1}^n T_{s_{t-1}s_t}^{a_{t-1}},\quad
  \P_{\!R}(r_{1:n}|s_{1:n}a_{1:n}) = \prod_{t=1}^n R_{s_{t-1}s_t}^{a_{t-1}r_t}
\eeqn
The probability of $\v r|\v a$ can be obtained by taking the product
and marginalizing $\v s$:
\bqan
  \P_{\!U}(r_{1:n}|a_{1:n})
  &=& \sum_{s_{1:n}}\P_{\!T}(s_{1:n}|a_{1:n}) \P_{\!R}(r_{1:n}|s_{1:n}a_{1:n})
\\[-1ex]
  &=& \sum_{s_{1:n}}\prod_{t=1}^n U_{s_{t-1}s_t}^{a_{t-1}r_t}
  \;=\; \sum_{s_n}[U^{a_0 r_1}\!\cdot\cdot\cdot U^{a_{n-1}r_n}]_{s_0s_n}
\eqan
where for each $a\in\A$ and $r'\in\R$, matrix
$U^{ar'}\in\SetR^{m\times m}$ is defined as $[U^{ar'}]_{ss'}\equiv
U_{ss'}^{ar'}:=T_{ss'}^a R_{ss'}^{ar'}$.
The right $n$-fold matrix product can be evaluated in time $O(m^2
n)$. This shows that $\v r$ given $\v a$ and $U$ can be
coded in $-\log\P_{\!U}$ bits.
The unknown $U$ needs to be estimated,
e.g.\ by the relative frequency $\hat
U_{ss'}^{ar'}:=n_{ss'}^{ar'}/n_{s\p}^{a\p}$.
Note that $\P_{\!U}$ completely ignores the observations $o_{1:n}$ and is
essentially independent of $\Phi$. Map $\Phi$ and hence $o_{1:n}$
enter $\P_{\!\hat U}$ (only and crucially) via the estimate $\hat U$.
The $M:=m(m-1)|\A|(|\R|-1)$ (independent) elements of $\hat U$ can
be coded to sufficient accuracy in $\fr12 M\log n$ bits, and $\Phi$
will be coded in $\CL(\Phi)$ bits. Together this leads to a code for
$\v r|\v a$ of length
\beq\label{ICost}
  \mbox{ICost}(\Phi|h_n) \;:=\; - \log\P_{\!\hat U}(r_{1:n}|a_{1:n}) + \fr12 M\log n + \CL(\Phi)
\eeq
In practice, $M$ can and should be chosen smaller like done in the
original \Cost\ function, and/or by using the restrictive model \req{rCodeEx}
for $R$, and/or by considering only non-zero frequencies \req{iidCodeLp}.
Analogous to \req{bestphi} we seek a $\Phi$ that minimizes ICost().

Since action evaluation is based on (discounted) reward sums, not
individual rewards, one may think of marginalizing $\P_{\!U}(\v r|\v
a,\Phi)$ even further, or coding rewards only approximately.
Unfortunately, the algorithms in Section \ref{secEE} that learn,
explore, and exploit MDPs require knowledge of the (exact)
individual rewards, so this improvement is not feasible.

\section{Discussion}\label{secDisc}

This section summarizes $\Phi$MDP, relates it to previous work, and
hints at more efficient incremental implementations and more
realistic {\em structured} MDPs (dynamic Bayesian networks).

\paradot{Summary}
Learning from rewards in general environments is an immensely
complex problem. In this paper I have developed a generic
reinforcement learning algorithm based on sound principles. The key
idea was to reduce general learning problems to finite state MDPs
for which efficient learning, exploration, and exploitation
algorithms exist. For this purpose I have developed a formal
criterion for evaluating and selecting good ``feature'' maps $\Phi$
from histories to states. One crucial property of $\Phi$MDP is that
it neither requires nor learns a model of the complete observation
space, but only for the reward-relevant observations as summarized
in the states.
The developed criterion has been inspired by MDL, which recommends
to select the (coding) model that minimizes the length of a suitable
code for the data at hand plus the complexity of the model itself.
The novel and tricky part in $\Phi$MDP was to deal with the states,
since they are not bare observations, but model-dependent processed
data. An improved Bayesian criterion, which integrates out the
states, has also been derived.
Finally, I presented a complete feature reinforcement learning
algorithm $\Phi$MDP-Agent(). The building blocks and computational
flow are depicted in the following diagram:

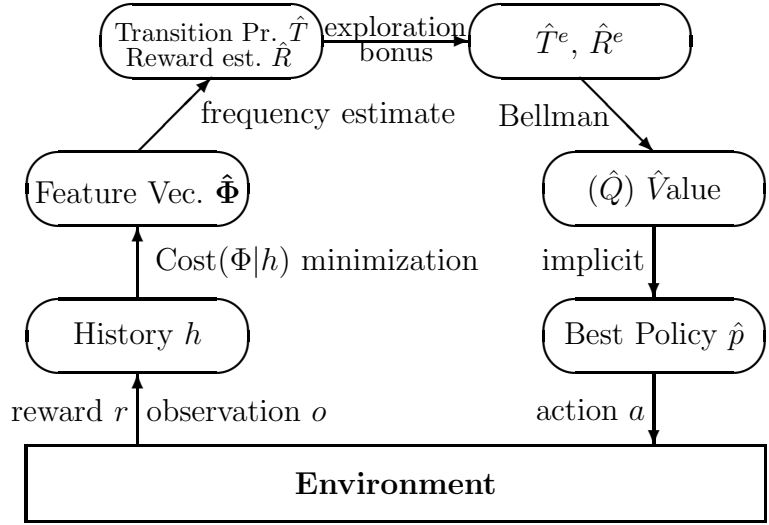
\begin{center}
\unitlength=2.7ex
\linethickness{0.4pt}
\begin{picture}(20,14)(0,0)
\thicklines
\put(0,0){\framebox(20,2)[cc]{\bf Environment}}
\put(3,5){\oval(6,2)\makebox(0,0)[cc]{History $h$}}
\put(3,9){\oval(6,2)\makebox(0,0)[cc]{Feature Vec.\ $\v{\hat\Phi}$}}
\put(5,13){\oval(6,2)\makebox(0,0)[cb]{\footnotesize Transition Pr. $\hat T$}
\makebox(0,0)[ct]{\footnotesize Reward est.\ $\hat R\quad$}}
\put(15,13){\oval(6,2)\makebox(0,0)[cc]{$\hat T^e$, $\hat R^e$}}
\put(17,9){\oval(6,2)\makebox(0,0)[cc]{($\hat Q$) $\hat V\!$alue}}
\put(17,5){\oval(6,2)\makebox(0,0)[cc]{Best Policy $\hat p$}}
\put(3,2){\vector(0,1){2}\makebox(0,2)[rc]{reward $r\;$}\makebox(0,2)[lc]{$\;$observation $o$}}
\put(3,6){\vector(0,1){2}\makebox(0,2)[lc]{$\;\;\Cost(\Phi|h)$ minimization}}
\put(3,10){\vector(1,1){2}\makebox(2,2)[lc]{$\!\!$frequency estimate}}
\put(8,13){\vector(1,0){4}\makebox(-4,0)[cb]{\small exploration}\makebox(-4,0)[ct]{\small bonus}}
\put(15,12){\vector(1,-1){2}\makebox(-1.2,-2)[rc]{Bellman}}
\put(17,8){\vector(0,-1){2}\makebox(0,-2)[rc]{implicit$\;$}}
\put(17,4){\vector(0,-1){2}\makebox(0,-2)[rc]{action $a\;$}}
\end{picture}
\end{center}

\paradot{Relation to previous work}
As already indicated here and there, $\Phi$MDP can be regarded as
extending the frontier of many previous important approaches to RL
and beyond:
{\em Partially Observable MDPs (POMDPs)} are a very important
generalization of MDPs \cite{Kaelbling:98}. Nature is still assumed
to be an MDP, but the states of nature are only partially observed
via some non-injective or probabilistic function. Even for finite
state space and known observation and transition functions, finding
and even only approximating the optimal action is (harder than NP)
hard \cite{Lusena:01,Madani:03}. Lifting any of the assumptions
causes conceptual problems, and when lifting more than one we enter
scientific terra nullius.
Assume a POMDP environment: POMDPs can formally (but not yet
practically) be reduced to MDPs over so-called (continuous) belief
states. Since $\Phi$MDP reduces every problem to an MDP, it is conceivable
that it reduces the POMDP to (an approximation of) its belief MDP.
This would be a profound relation between $\Phi$MDP and POMDP, likely
leading to valuable insights into $\Phi$MDP and proper algorithms for
learning POMDPs. It may also help us to restrict the space of
potentially interesting features $\Phi$.
{\em Predictive State Representations (PSRs)} are very interesting,
but to this date in an even less developed stage \cite{Singh:03}
than POMDPs.
{\em Universal AI} \cite{Hutter:04uaibook} is able to
optimally deal with arbitrary environments, but the resulting AIXI
agent is computationally intractable \cite{Hutter:07aixigentle} and
hard to approximate \cite{Pankov:08,Hutter:06aixifoe}.
{\em Bayesian RL} algorithms
\cite{Dearden:99,Duff:02,Poupart:06,Ross:08bayes} (see also
\cite[Chp.11]{Kumar:86}) can be regarded as implementations of
the AI$\xi$ models \cite{Hutter:06aixifoe}, which are down-scaled
versions of AIXI, but the enormous computational demand still
severely limits this approach.
$\Phi$MDP essentially differs from ``generative'' Bayesian RL and AI$\xi$
in that it neither requires to specify nor to learn a (complete
stochastic) observation model. It is a more ``discriminative''
approach \cite{Liang:08}. Since $\Phi$MDP ``automatically'' models only
the relevant aspects of the environment, it should be
computationally less demanding than full Bayesian RL.
{\em State aggregation} methods have been suggested earlier for
solving large-scale MDP planning problems by grouping (partitioning)
similar states together, resulting in (much) smaller block MDPs
\cite{Givan:03}. But the bi-simulation metrics used require
knowledge of the MDP transition probabilities. $\Phi$MDP might be regarded
as an approach that lifts this assumption.
{\em Suffix trees} \cite{McCallum:96} are a simple class
of features $\Phi$. $\Phi$MDP combined with a local search
function that expands and deletes leaf nodes is closely
related to McCallum's U-Tree algorithm \cite{McCallum:96},
with a related but likely different split\&merge criterion.
{\em Miscellaneous}: $\Phi$MDP also extends
the theory of model selection (e.g.\ MDL \cite{Gruenwald:07book})
from passive to active learning.

\paradot{Incremental updates}
As discussed in Section \ref{secCM}, most search algorithms are
local in the sense that they produce a chain of ``slightly''
modified candidate solutions, here $\Phi$. This suggests a potential
speedup by computing quantities of interest incrementally, which
becomes even more important in the $\Phi$DBN case
\cite{Hutter:09phidbn,Hutter:09phidbnx}.

Computing $\Cost(\Phi)$ takes at most time $O(|\S|^2|\A||\R|)$. If
we split or merge two states, we can incrementally update the cost
in time $O(|\S||\A||\R|)$, rather than computing it again from
scratch. In practice, many transition $T_{ss'}^a$ don't occur, and
$\Cost(\Phi)$ can actually be computed much faster in time
$O(|\{n_{ss'}^{ar}>0\}|)$, and incrementally even faster.

Iteration algorithms for \req{BellmanEq} need an initial value for
$V$ or $Q$. We can take the estimate $\hat V$ from a previous $\Phi$
as an initial value for the new $\Phi$. For a merge operation we can
average the value of both states, for a split operation we could give
them the same initial value. A significant further speedup can be
obtained by using prioritized iteration algorithms that concentrate
their time on badly estimated states, which are in our case
(states close to) the new ones \cite{Sutton:98}.

Similarly, results from cycle $n$ can be (re)used for the next cycle
$n+1$. For instance, $\hat V$ can simply be reused as an initial
value in the Bellman equations, and ICost$(\Phi)$ can be updated in
time $O(|\S|^2)$ or even faster if $U$ is sparse.

\paradot{Feature dynamic Bayesian networks} 
The use of ``unstructured'' MDPs, even our $\Phi$-optimal ones, is
clearly limited to very simple tasks. Real world problems are
structured and can often be represented by dynamic Bayesian networks
(DBNs) with a reasonable number of nodes.
Our $\Phi$ selection principle can be adapted from MDPs to the
conceptually much more complex DBN case. The primary purpose of this
Part I was to explain the key concepts on an as simple model as
possible, namely unstructured finite MDPs, to set the stage for
developing the more realistic $\Phi$DBN in Part II
\cite{Hutter:09phidbnx}.

\paradot{Outlook}
The major open problems are to develop smart $\Phi$ generation and
smart stochastic search algorithms for $\Phi^{best}$, and to
determine whether minimizing \req{ICost} is the right criterion.

\paradot{Acknowledgements}
My thanks go to Pedro Ortega, Sergey Pankov, Scott Sanner, J\"urgen
Schmidhuber, and Hanna Suominen for feedback on earlier drafts.

\addcontentsline{toc}{section}{\refname}

\begin{small}
\newcommand{\etalchar}[1]{$^{#1}$}

\end{small}

\newpage\appendix
\section{List of Notation}\label{secApp}

\begin{tabbing}
  \hspace{2.3cm} \= \hspace{11cm} \= \kill
  {\bf Interface structures} \>                                                              \\[0.5ex]
  $\O$               \> = finite or infinite set of possible observations                    \\[0.5ex]
  $\A$               \> = (small) finite set of actions                                      \\[0.5ex]
  $\R$               \> = $\{0,1\}$ or $[0,R_{max}]$ or other set of rewards                 \\[0.5ex]
  $n\in\SetN$        \> = current time                                                       \\[0.5ex]
  $o_t r_t a_t$      \> = $ora_t\in\O\times\R\times\A$ = true observation, reward, action at time $t$ \\[2ex]
  {\bf Internal structures for $\mathbf\Phi$MDP} \>                                          \\[0.5ex]
  $\log$             \> = binary logarithm                                                   \\[0.5ex]
  $t\in\{1,...,n\}$  \> = any time                                                           \\[0.5ex]
  $i\in\{1,...,m\}$  \> = any state index                                                    \\[0.5ex]
  $\v x=x_{1:n}$     \> = $x_1...x_n$ (any $x$)                                              \\[0.5ex]
  $x_\p,x_*,\v x_\vi$\> = $\sum_j x_j$, $\bigcup_j x_j$, $(x_1,...,x_l)$ (any $x,j,l$)       \\[0.5ex]
  $\hat X$           \> = estimate of $X$ (any $X$)                                          \\[0.5ex]
  $\H$               \> = $(\O\times\R\times\A)^*\times\O\times\R$ = possible histories              \\[0.5ex]
  $h_n$              \> = $ora_{1:n-1}o_n r_n$ = actual history at time $n$                      \\[0.5ex]
  $\S$               \> = $\{s^1,...,s^m\}$ = internal finite state space (can vary with $n$)\\[0.5ex]
  $\Phi:\H\to\S$     \> = state or feature summary of history                                \\[0.5ex]
  $s_t$              \> = $\Phi(h_t)\in\S$ = realized state at time $t$                      \\[0.5ex]
  $\P(\cdot)$        \> = probability over states and rewards or parts thereof               \\[0.5ex]
  $\CL(\cdot)$       \> = code length                                                        \\[0.5ex]
  MDP                \> = $(\S,\A,T,R)$ = Markov Decision Process                            \\[0.5ex]
  $T_{ss'}^a$        \> = $\P(s_t=s'|s_{t-1}=s,a_{t-1}=a)$ = transition matrix               \\[0.5ex]
  $s\stackrel{a}\to s'(r')$ \> = action $a$ in state $s$ resulted in state $s'$ (and reward $r'$) \\[0.5ex]
  $\T_{ss'}^{ar'}$   \> = set of times $t\in\{1,...,n\}$ at which $s\stackrel{a}\to s'r'$    \\[0.5ex]
  $n_{ss'}^{ar'}$    \> = $|\T_{ss'}^{ar'}|$ = number of times $t\in\{1,...,n\}$ at which $s\stackrel{a}\to s'r'$ \\[0.5ex]
  $\Cost(\Phi|h)$    \> = cost (evaluation function) of $\Phi$ based on history $h$             \\[0.5ex]
  ICost$(\Phi|h)$    \> = improved cost function                                             \\[0.5ex]
  $Q_s^{*a},V_s^*$   \> = optimal ($Q$) $V\!$alue (of action $a$) in state $s$                 \\[0.5ex]
  $\g\in[0;1)$       \> = discount factor ($(1-\g)^{-1}$ is effective horizon)               \\[0.5ex]
\end{tabbing}

\end{document}